\pdfoutput=1

\documentclass[11pt]{article}

\usepackage{ACL2023}

\usepackage{amsmath}
\usepackage{times}
\usepackage{latexsym}
\usepackage{multirow}
\usepackage{amssymb}
\usepackage{graphicx}
\usepackage{arydshln}
\usepackage{hyperref}
\usepackage[T1]{fontenc}

\usepackage[utf8]{inputenc}

\usepackage{microtype}

\usepackage{inconsolata}
\usepackage{xspace}
\title{Modality Adaption or Regularization?\\A Case Study on End-to-End Speech Translation}

\author{Yuchen Han\textsuperscript{1}, Chen Xu\textsuperscript{1}, Tong Xiao\textsuperscript{1,2}\thanks{\xspace\xspace Corresponding author.}, Jingbo Zhu\textsuperscript{1,2}\\
\textsuperscript{1}School of Computer Science and Engineering, Northeastern University, Shenyang, China\\
\textsuperscript{2}NiuTrans Research, Shenyang, China\\
\texttt{hanyuchen114@gmail.com,xuchennlp@outlook.com}\\
\texttt{\{xiaotong,zhujingbo\}@mail.neu.edu.cn}}

\begin{document}
\maketitle
\begin{abstract}
Pre-training and fine-tuning is a paradigm for alleviating the data scarcity problem in end-to-end speech translation (E2E ST). The commonplace ``modality gap'' between speech and text data often leads to inconsistent inputs between pre-training and fine-tuning. However, we observe that this gap occurs in the early stages of fine-tuning, but does not have a major impact on the final performance. On the other hand, we find that there has another gap, which we call the ``capacity gap'': high resource tasks (such as ASR and MT) always require a large model to fit, when the model is reused for a low resource task (E2E ST), it will get a sub-optimal performance due to the over-fitting. In a case study, we find that the regularization plays a more important role than the well-designed modality adaption method, which achieves 29.0 for en-de and 40.3 for en-fr on the MuST-C dataset. Code and models are available at \href{https://github.com/hannlp/TAB}{https://github.com/hannlp/TAB}.
\end{abstract}

\section{Introduction}
End-to-end speech translation (E2E ST) employs a direct model to translate source language speech into target language text, which has low latency and can avoid the ``error propagation'' problem in traditional cascade methods \citep{DBLP:conf/interspeech/WeissCJWC17}. However, compared to automatic speech recognition (ASR) and machine translation (MT) models used in cascade methods, E2E ST models typically have limited training data \citep{DBLP:journals/csl/CattoniGBNT21}, which can result in sub-optimal performance.

\begin{figure}
    \centering
    \includegraphics[width=75mm]{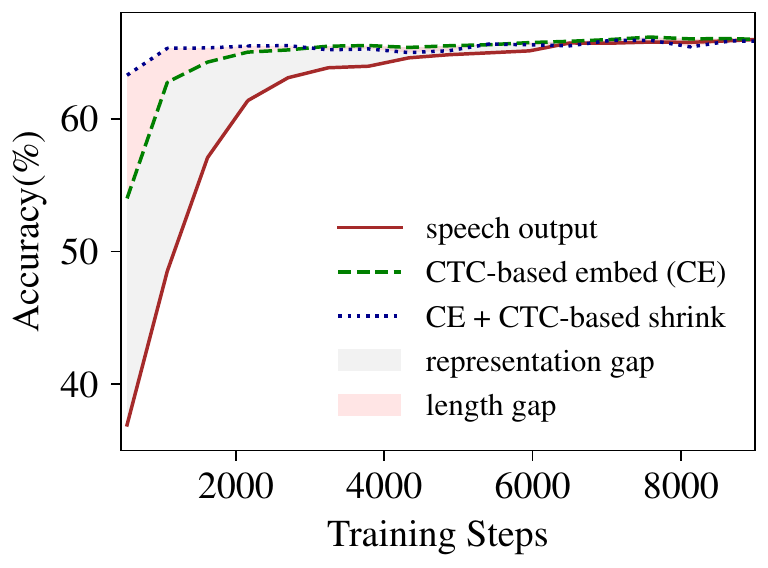}
    \caption{The modality gap in the PT-FT paradigm. We decoupled the encoder to speech encoder and text encoder \citep{DBLP:conf/aaai/WangWLY020, DBLP:conf/acl/XuHLZHJXZ20}, and compared the accuracy curve of fine-tuning for different inputs of the text encoder. We find that the modality gap only exists at the initial stage of the fine-tuning.}
    \label{fig:modality-gap}
\end{figure}

Transferring the knowledge from the related tasks (e.g. ASR and MT) is a widely-used approach for E2E ST to achieve optimal performance \citep{DBLP:conf/acl/TangPLWG20,DBLP:journals/corr/abs-2212-01778/auxdata}. However, the difference between tasks and data makes the transfer process more challenging \citep{DBLP:conf/aaai/WangWLY020}. The inconsistency of length and representation between speech and text leads to the ``modality gap'' \citep{DBLP:journals/corr/abs-2010-14920/bridgegap}, which exists in scenarios where the inputs of the model change, such as in the pre-training and fine-tuning (PT-FT) paradigm \citep{DBLP:conf/acl/XuHLZHJXZ20} or in the multi-task learning (MTL) methods \citep{DBLP:conf/interspeech/YeW021}. Thus, the connectionist temporal classification (CTC) \citep{DBLP:conf/icml/GravesFGS06} based adapters \citep{DBLP:journals/corr/abs-2010-14920/bridgegap,DBLP:conf/acl/XuHLZHJXZ20} have been proposed to transform the original speech output into a text-like sequence. Recently, consistency training methods have achieved promising results by using a better branch, such as the mix-up branch \citep{DBLP:conf/acl/FangYLFW22} or the text branch \citep{DBLP:journals/corr/abs-2212-03657/m3st}, to promote cross-modal learning and support the original speech output. However, we find that the ``modality gap'' does not exist throughout the training process in Figure~\ref{fig:modality-gap}. While consistency training methods have some regularization function \citep{DBLP:conf/nips/LiangWLWMQCZL21,DBLP:conf/acl/GuoMZ022} to help the model overcome over-fitting and be fully trained. A natural question arises: \textit{
Are modality adaption methods always effective when an E2E ST model is fully trained?}

In this work, we aim to investigate how much of the improvement is due to the modality adaption or regularization methods. To achieve this, we adopt the PT-FT and encoder decouple paradigm and establish a framework that incorporates adjustable modality adaption and consistency training methods. Through extensive experiments on the MuST-C en-de and en-fr benchmarks, we observe that:

\begin{itemize}
    \item The modality adaption method in PT-FT only accelerates the early phase of fine-tuning, but does not provide a significant improvement for a fully trained model.
    \item We obtained 29.0 and 40.3 on the MuST-C en-de and en-fr datasets, but regularization played a major role, which confirming that the ``capacity gap'' is more severe than the ``modality-gap'' in E2E ST.
\end{itemize}

\section{Our Case Study: TAB}
\begin{figure*}
    \centering    \includegraphics[width=168mm]{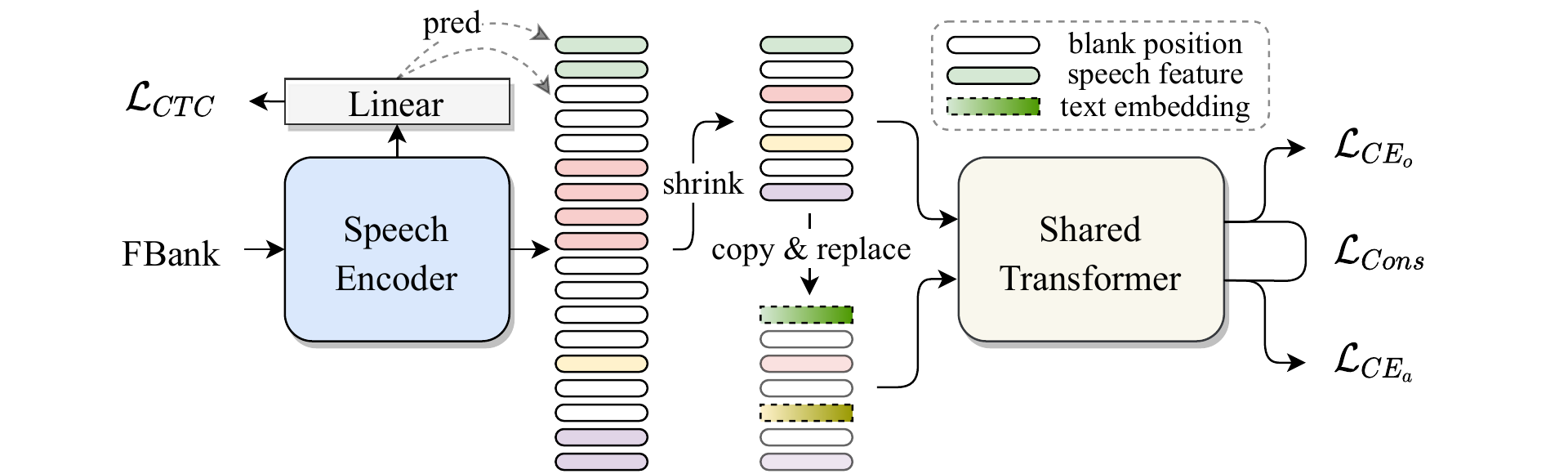}
    \caption{An illustration of TAB. The speech output is first shrunk based on the CTC predicted path, then copied and replaced by text embedding with a probability $p^*$ to obtain the auxiliary branch. Two branches are fed into a shared transformer, and the final predictions are kept consistent by consistency loss.}
    \label{fig:method}
\end{figure*}

\subsection{Architecture}
\label{Architecture}
The E2E ST corpus $\mathcal{D_{ST}}$ usually consists of three parts: the source language speech $\boldsymbol{s}$, the corresponding transcripts $\boldsymbol{x}$, and the target translation $\boldsymbol{y}$. The overall framework, as shown in Figure~\ref{fig:method}, consists of a speech encoder and a shared transformer.

\noindent\textbf{Speech Encoder.} The speech encoder encodes the source language speech $\boldsymbol{s}$ into speech output $\boldsymbol{h}$:
\begin{eqnarray}
 \boldsymbol{h} & = & \mathrm{ENC_{speech}}(\boldsymbol{s};\theta_s)
\end{eqnarray}

We employ the CTC loss at the top of the speech encoder \citep{DBLP:conf/aaai/WangWLY020, DBLP:journals/corr/abs-2010-14920/bridgegap, DBLP:conf/acl/XuHLZHJXZ20} to predict an alignment path $\boldsymbol{\pi}$ by speech output $\boldsymbol{h}$ based on a conditionally independent assumption $p(\boldsymbol{\pi}|\boldsymbol{h})=\prod\nolimits_t^{|\boldsymbol{h}|} p(\pi_t|\boldsymbol{h_t})$:
\begin{eqnarray}
p(\pi_t|\boldsymbol{h_t})&=&\textrm{Softmax}(\textrm{Linear}(\boldsymbol{h_t}; \theta_{ctc}))
\end{eqnarray}
where $\pi_t\in\mathcal{V}^+$, which is an extended source language vocabulary by introducing a ``blank'' token. The path $\boldsymbol{\pi}$ can be mapped to the transcript $\boldsymbol{x}$ by removing all repeated labels and blanks, such an operation is called $\beta$. The CTC loss is defined as the negative log probability of all possible alignment paths $\beta^{-1}(\boldsymbol{x})$ between $\boldsymbol{h}$ and $\boldsymbol{x}$:
\begin{eqnarray}
\mathcal{L}_{CTC}&=&-\sum\nolimits_{\boldsymbol{\pi}\in\beta^{-1}(\boldsymbol{x})}\textrm{log}p(\boldsymbol{\pi}|\boldsymbol{h})
\end{eqnarray}

\noindent\textbf{Shared Transformer.} The shared transformer accepts two inputs: the original branch $\boldsymbol{o}$ and the auxiliary branch $\boldsymbol{a}$, both of equal length. We leave the details of these to the Section~\ref{Tuning with Auxiliary Branch (TAB)}. The encoder and decoder of the shared transformer are utilized to obtain the final predictions $\mathcal{P}_j=p(y_j|\boldsymbol{y_{<j}, o}; \theta_t)$ and $\mathcal{Q}_j=p(y_j|\boldsymbol{y_{<j}, a}; \theta_t)$, respectively. The cross-entropy losses are then calculated as follows:
\begin{eqnarray}
\mathcal{L}_{CE_o}&=&-\sum\nolimits_{j=1}^{|y|}\textrm{log} \mathcal{P}_j \\
\mathcal{L}_{CE_a}&=&-\sum\nolimits_{j=1}^{|y|}\textrm{log} \mathcal{Q}_j
\end{eqnarray}

\subsection{Tuning with Auxiliary Branch (TAB)}
\label{Tuning with Auxiliary Branch (TAB)}
The speech encoder and shared transformer are initialized with pre-trained ASR and MT models. During fine-tuning, we aim to build a text-like auxiliary branch which includes some textual representations like \citet{DBLP:conf/acl/FangYLFW22,DBLP:conf/emnlp/ZhangZA0D0W22} for modality adaption, and provide an adjustable probability to control the degree of it. To obtain the textual embedding, we utilize the ability of the CTC alignment, where $\pi_t=\textrm{argmax}(p(\pi_t|\boldsymbol{h_t}))$ is an id of $\mathcal{V}^+$ that denotes the corresponding CTC-predicted token of the speech feature $\boldsymbol{h_t}$.

\noindent\textbf{Shrink.} To eliminate the effect of too many ``blank'' positions in the sequence, we first average the consecutively repeated features (e.g. $\pi_i=...=\pi_j=c_{i\to j}$) in the speech output $\boldsymbol{h}$ to obtain the original branch $\boldsymbol{o}$, where $\boldsymbol{o_k} = (\boldsymbol{h_i}+...+\boldsymbol{h_j})\cdot \frac{1}{j-i}$.

\noindent\textbf{Copy \& Replace.} We copy $\boldsymbol{o}$ to a new sequence $\boldsymbol{a}$ to ensure that the auxiliary branch has the same length as the original branch. Each position $\boldsymbol{a_k}$ in the new sequence is then replaced with its CTC predicted embedding $\mathrm{Embedding}(c_{i\to j})$ with a probability $p^*$ if $c_{i\to j}$ is not a ``blank''. Here, $\mathrm{Embedding}$ is an embedding matrix initialized by the pre-trained MT model. The replacement probability $p^*$ can be adjusted to control the degree of modality adaption, which can be a fixed or a dynamic value, as discussed in Section~\ref{sec:help}. It is important to note that the auxiliary branch provides a regularization function due to the replacement operation or dropout \citep{DBLP:conf/nips/LiangWLWMQCZL21}. This effect will be further discussed in Section~\ref{sec:always}. 

\noindent\textbf{Fine-tuning strategy.} To utilize the auxiliary branch, a consistency loss is introduced to enforce consistency between two output distributions:

\begin{eqnarray}
\mathcal{L}_{Cons}&=&\sum\nolimits_{j=1}^{|y|}\mathcal{D}(\mathcal{P}_j,\mathcal{Q}_j)
\end{eqnarray}
where $\mathcal{D}$ denotes the loss term. The final loss used in TAB is formulated as follows:
\begin{eqnarray}
\mathcal{L}\;=\;\mathcal{L}_{CE_o}+\mathcal{L}_{CE_a}+\lambda\mathcal{L}_{CTC}+\alpha\mathcal{L}_{Cons}
\end{eqnarray}

\section{Experimental Setup}
\noindent\textbf{Datasets and Pre-processing.} We conducted our experiments on the  MuST-C \citep{DBLP:conf/interspeech/GulatiQCPZYHWZW20} dataset for two language directions: En-De and En-Fr. The dev set was used for validation, and the tst-COMMON set was used for reporting our results. For training the English ASR model, we used the LibriSpeech \citep{DBLP:conf/icassp/PanayotovCPK15} dataset. The WMT16 En-De and WMT14 En-Fr datasets were used to train the MT models. Table~\ref{tab:datastat}  presents the statistics of all the datasets used in our pre-training and fine-tuning processes.

\begin{table}[h]
\centering
\begin{tabular}{l|ccc}
\hline
Dataset & ASR(H) & MT(S) & ST(H/S)\\
\hline
En-De & 960 & 4.5M & 408/234K \\
En-Fr & 960 & 36M & 492/280K \\
\hline
\end{tabular}
\caption{Statistics of ASR, MT and ST data for two language pairs, where H denotes hours and S denotes sentences.}
\label{tab:datastat}
\end{table}

We preprocessed the speech input by extracting 80-dimensional log-mel filterbank features and removing utterances with more than 3,000 frames. The vocabulary, which has a size of 10k, is shared between the source and target languages and was trained using the SentencePiece \citep{DBLP:conf/emnlp/KudoR18} model from the MuST-C dataset. 

\noindent\textbf{Model Configuration.}
All experiments were implemented using the fairseq toolkit \citep{DBLP:conf/naacl/fairseq-OttEBFGNGA19}. Two convolutional layers with a stride of 2 were introduced to downsample the input speech features. We used the Conformer \citep{DBLP:conf/interspeech/GulatiQCPZYHWZW20} as our speech encoder, which consists of 12 layers. Both the text encoder and decoder in the shared transformer have 6 layers. Each layer in our model has 512 hidden units, 2048 feed-forward size, and 8 attention heads. The ASR and MT models were pre-trained with external data and fine-tuned with the MuST-C dataset.

\noindent\textbf{Training and Inference.} We used the Adam optimizer with $\beta_1=0.9$ and $\beta_2=0.997$ in MT, while $\beta_2=0.98$ in ASR and ST, respectively. During ASR pre-training, each batch has up to 800k frames, and the learning rate and warmup were 1.4e-3 and 10000. During MT pre-training, each batch has up to 33k tokens, and the learning rate and warmup were 1e-3 and 8000. During ST fine-tuning, each batch has up to 320k frames, and the learning rate and warmup were 7e-4 and 4000. The hyper-parameter $\lambda$ was set to 0.3 for both pre-training and fine-tuning. We used dropout with a ratio of 0.1 during pre-training and 0.15 during fine-tuning, and label smoothing with a value of 0.1. All training was stopped early if the loss (ASR and MT) or BLEU (E2E ST) on the dev set did not improve for twenty epochs. During inference, we averaged the model parameters of the best 10 checkpoints for evaluation. We used a beam search with a beam size of 5 for all models. We reported the case-sensitive SacreBLEU \citep{DBLP:conf/wmt/Post18}. All models were trained on 4 Titan RTX GPUs.

\section{Results and Discussion}
\subsection{Which type of consistency loss is best?}
\begin{table}
\centering
\resizebox{\linewidth}{!}{
\begin{tabular}{lccc} 
\hline
\multirow{2}{*}{Loss term} & \multicolumn{3}{c}{BLEU} \\
\cline{2-4} & dev & tst-COMMON & avg. \\
\hline
$\textrm{None ($\alpha=0$)}$ & 27.69 & 27.99 & 27.84 \\
\cdashline{1-4}
$\textrm{JSD}$ & 27.76 & 28.49 & 28.13 \\
$\textrm{KL}_{orig\to aux}$  & 28.47 & 28.49 & 28.48 \\
$\textrm{KL}_{aux\to orig}$  & 28.26 & 28.78 & 28.52 \\
$\textrm{bi-KL}$ & 28.43 & 28.78 & \textbf{28.61} \\
\hline
\end{tabular}
}
\caption{BLEU scores on MuST-C en-de dev and tst-COMMON sets with different consistency loss terms.}
\label{tab:lossterm}
\end{table}
The choice of an appropriate consistency loss is crucial in leveraging the knowledge from the auxiliary branch, whether it is due to modality adaption or regularization. We conducted experiments with different loss terms with $\alpha=1$ and $p^*=0.2$. As shown in Table~\ref{tab:lossterm}, the results indicate that a consistency loss is necessary to improve performance. The Jensen-Shannon Divergence (JSD) loss and the unidirectional-KL loss were found to be slightly worse than the bidirectional-KL (bi-KL) loss. Therefore, we selected the bi-KL loss for the subsequent experiments.

\subsection{Whether \& when the modality gap exist?}
\label{sec:ifandwhen}
\begin{figure}
    \centering
    \includegraphics[width=75mm]{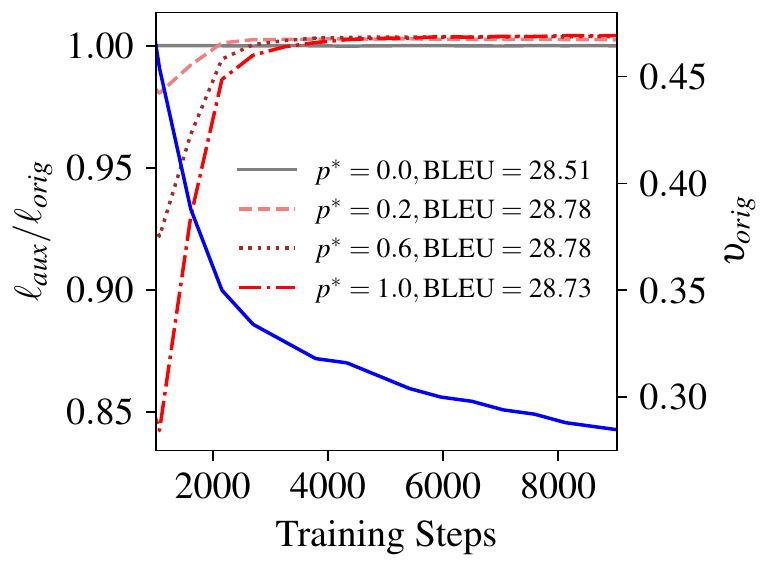}
    \caption{The ratios of the auxiliary branch loss to the original branch loss $\ell_{aux}/\ell_{orig}$ at each training step under different fixed values of $p^*$. Additionally, we show the uncertainty of the original branch $\upsilon_{orig}$ (represented by the blue line) at the same time.}
    \label{fig:mix-trans}
\end{figure}
In Figure~\ref{fig:mix-trans}, we present the $\ell_{aux}/\ell_{orig}$ curve of fine-tuning, which represents the ratio of the auxiliary branch loss $\ell_{aux}$ to the original branch loss $\ell_{orig}$. The only difference between the two branches is the replaced textual embeddings for modality adaption in the auxiliary branch. We investigate the effect of this operation during fine-tuning under different replacement probabilities.

Our results show that, in the early fine-tuning phase, the $\ell_{aux}$ always lower than $\ell_{orig}$ when $p^*>0$, indicating that the model expects some textual representations like the input in pre-training. However, in the later fine-tuning process, the $\ell_{aux}$ is slightly higher than the $\ell_{orig}$, suggesting that the model starts to get confused about replacement operations due to the introduction of some kind of noise by the destruction of the original sequence. Moreover, we find that the maximum $p^*=1.0$ always has the lowest ratio at the beginning and the highest ratio at other times. This confirms that the modality gap exists but not throughout the entire fine-tuning process. 

\subsection{Does the modality adaption help?}
\label{sec:help}
We experimented with different fixed replacement ratios $p^*$ over the range $[0.0,0.2,0.6,1.0]$ under $\alpha=1.0$ in Figure~\ref{fig:mix-trans}. Our results showed that the method with modality adaption $p^*>0$ consistently outperformed $p^*=0$. Furthermore, as observed in Section~\ref{sec:ifandwhen}, there exists a ``tipping point'' in the fine-tuning process where the modality gap disappears. Before this point, we should use a higher $p^*$, while a lower $p^*$ should be more effective after this point, rather than using a fixed value. We discovered that the uncertainty of the original branch $\upsilon_{orig}$, which is defined by the normalized entropy of $P_j$ \footnote{More precise definitions are $P_j(t)$ and $\upsilon_{orig}(t)$, with the symbol of the training step ``t'' omitted for brevity.}, is strongly related to this point, as shown in Figure~\ref{fig:mix-trans}:
\begin{eqnarray}
v_{orig}&=&-\frac{1}{\textrm{log}V}\cdot\frac{1}{|y|}\sum\nolimits_{j=1}^{|y|}\mathcal{P}_j \textrm{log} \mathcal{P}_j
\end{eqnarray}
where $V$ is the vocabulary size. We then proposed a dynamic replacement probability derived from $\upsilon_{orig}$ at each step: $p^*=\gamma\cdot\upsilon_{orig}$, where $\gamma$ is a hyper-parameter set to 0.5 in all experiments. When we use a dynamic replacement ratio of $p^*=0.5\cdot\upsilon_{orig}$, we denote it as $p^*=\upsilon$. By adopting this dynamic replacement ratio, we achieved a BLEU score of 28.87 on the MuST-C en-de dataset.
\subsection{Is modality adaption always effective?}
\label{sec:always}
\begin{figure}
    \centering
    \includegraphics[width=75mm]{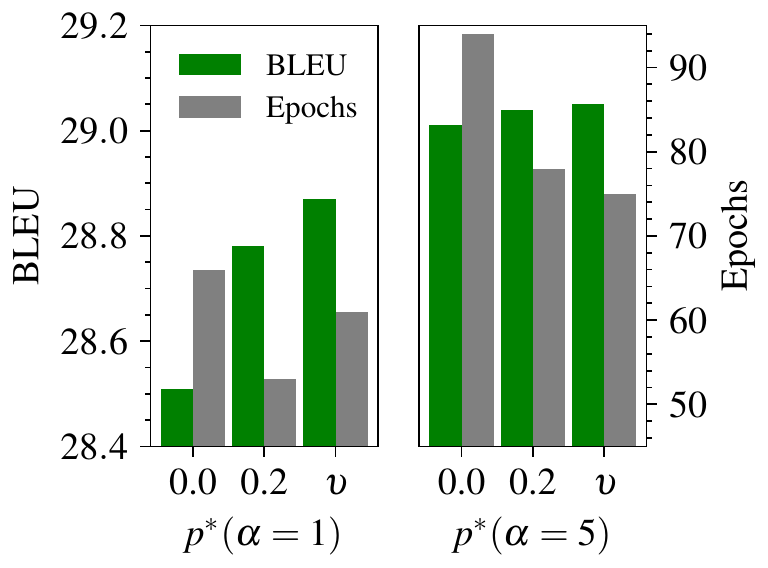}
    \caption{BLEU scores and the epochs required for convergence with different values of $\alpha$ and $p^*$ on the MuST-C en-de dataset. The $\upsilon$ denotes the use of an uncertainty-based ($\gamma=0.5$) dynamic value for $p^*$.}
    \label{fig:diffalpha}
\end{figure}
The consistency loss has been demonstrated to have a regularization effect on two branches with different outputs \citep{DBLP:conf/nips/LiangWLWMQCZL21,DBLP:conf/acl/GuoMZ022}. When $p^*=0$, there is no modality adaption through the replacement operation, but the dropout still causes different outputs for the two branches, the TAB degenerates into a pure regularization method. The hyper-parameter $\alpha$ can be used to control the degree of regularization, where a higher $\alpha$ indicates stronger regularization. By increasing $\alpha$ to 5, it is observed that the gap between the modality adaption method ($p^*=0.2$ or $p^*=\upsilon$) and the pure regularization method ($p^*=0$) decreases (29.05 vs 29.01) in Figure~\ref{fig:diffalpha}, although the pure regularization method always required more epochs to converge, which can be attributed to more complete training. These findings confirm that the modality adaption method on PT-FT can accelerate the early phase of fine-tuning, but does not significantly enhance the final performance when a sufficient level of regularization has been applied to the model. This also highlights the issue of over-fitting being a more serious problem in E2E ST than the ``modality gap'', and that better regularization and a longer time for fine-tuning can help to eliminate the ``modality gap'' problem in the PT-FT paradigm.

\subsection{Final results and comparison of methods}
\begin{table}
\centering
\resizebox{\linewidth}{!}{
\begin{tabular}{lcc} 
\hline
\multirow{2}{*}{Methods} & \multicolumn{2}{c}{BLEU} \\
\cline{2-3} & En-De & En-Fr \\ 
\hline
SATE \citep{DBLP:conf/acl/XuHLZHJXZ20} & 28.1 & - \\
STEMM$^\dagger$ \citep{DBLP:conf/acl/FangYLFW22} & 28.7 & 37.4 \\
ConST$^\dagger$ \citep{DBLP:conf/naacl/YeWL22} & 28.3 & 38.3 \\
M$^3$ST$^{\dagger\dagger}$ \citep{DBLP:journals/corr/abs-2212-03657/m3st} & 29.3 & 38.5 \\
WACO$^\dagger$ \citep{DBLP:journals/corr/abs-2212-09359/waco} & 28.1 & 38.1 \\
AdaTrans \citep{DBLP:journals/corr/abs-2212-08911/adatrans} & 28.7 & 38.7 \\
CRESS$^\dagger$ \citep{DBLP:journals/corr/abs-2305-08706/cress} & 28.9 & -\\
CRESS$^{\dagger\dagger}$ \citep{DBLP:journals/corr/abs-2305-08706/cress} & $\mathbf{29.4}$ & 40.1 \\
\hline
baseline (CTC) & 27.9 & 39.1 \\
TAB ($p^*=\upsilon, \alpha=1$) & 28.9 & 39.8 \\
TAB ($p^*=0, \alpha=5$) & $\mathbf{29.0}$ & 39.9 \\
\cdashline{1-3}
TAB ($p^*=\upsilon, \alpha=5$) & $\mathbf{29.0}$ & $\mathbf{40.3}$ \\
\hline
\end{tabular}
}
\caption{BLEU scores on the MuST-C tst-COMMON set. Methods marked with $\dagger$ employ self-supervised pre-training models that were trained on unsupervised speech data instead of ASR data. $\dagger$ refers to Wav2Vec2.0 and $\dagger\dagger$ refers to HuBERT.}
\label{tab:mainexp}
\end{table}
The results for the MuST-C dataset are shown in Table~\ref{tab:mainexp}. The modality adaption method shows an improvement of 1.0 and 0.7 BLEU points over our CTC baseline at a lower consistency level ($\alpha=1$). However, the pure regularization method with $\alpha=5$ slightly outperforms it ($+0.1$ BLEU), and outperforms all other methods designed for modality adaption ($+0.1\sim1.2$ BLEU), except those using HuBERT as the feature extractor. When we combined our modality adaption method with a higher level of consistency ($\alpha=5$), further improvement can still be achieved, but not consistently in all languages. Our hypothesis is that the replacement operation in TAB not only releases the modality gap in the early fine-tuning phase but also introduces noise in the later stages. This noise can bring better regularization performance \citep{DBLP:conf/acl/GuoMZ022,DBLP:conf/naacl/GaoH0022} when a higher consistency level is given. In general, regularization brings more improvement in TAB than modality adaption, and a better regularization is helpful for E2E ST models to be fully trained.

\section{Conclusion}
Through a case study, we have demonstrated that the ``modality gap'' in the PT-FT paradigm for E2E ST is only present in the early stages of fine-tuning. Although a modality adaption method can accelerate the convergence speed, it does not significantly improve the final performance of a fully trained model. However, the over-fitting and ``capacity gap'' are more critical issues in E2E ST, and a good regularization technique can help in fully training the model.

\section*{Acknowledgement}
The authors would like to thank anonymous reviewers for their insightful comments.
This work was supported in part by the National Science Foundation of China (No. 62276056), the National Key R\&D Program of China, the China HTRD Center Project (No. 2020AAA0107904), the Natural Science Foundation of Liaoning Province of China (2022-KF-16-01), the Yunnan Provincial Major Science and Technology Special Plan Projects (No. 202103AA080015), the Fundamental Research Funds for the Central Universities (Nos. N2216016, N2216001, and N2216002), and the Program of Introducing Talents of Discipline to Universities, Plan 111 (No. B16009).

\section*{Limitations}
Although our work has achieved high performance, there are still some limitations that need to be addressed in future work:
\begin{itemize}
    \item Our work was only carried out under the assumption that there is sufficient ASR and MT data, and relied on transcriptions for CTC loss to perform alignment and predict. This assumption may not hold in real-world scenarios, and we plan to investigate the performance of our approach under more diverse data conditions in the future.
    \item We only attempted to feed two branches into the shared transformer in order to ensure a fair comparison between the pure regularization method and the modality adaption method. However, this approach may have resulted in sub-optimal regularization performance compared to methods that feed all branches into the whole model, as demonstrated by \citet{DBLP:conf/nips/LiangWLWMQCZL21, DBLP:conf/acl/GuoMZ022, DBLP:conf/naacl/GaoH0022}.
\end{itemize}

\bibliography{anthology,custom}
\bibliographystyle{acl_natbib}

\appendix

\end{document}